\def\year{2020}\relax
\documentclass[letterpaper]{article} % DO NOT CHANGE THIS
\usepackage{aaai20}  % DO NOT CHANGE THIS
\usepackage{times}  % DO NOT CHANGE THIS
\usepackage{helvet} % DO NOT CHANGE THIS
\usepackage{courier}  % DO NOT CHANGE THIS
\usepackage[hyphens]{url}  % DO NOT CHANGE THIS
\usepackage{graphicx} % DO NOT CHANGE THIS
\usepackage{graphicx}  % DO NOT CHANGE THIS
\usepackage{amsmath,amsfonts,amssymb,algorithm,algorithmic}
\usepackage{multirow}
\usepackage[export]{adjustbox}
\usepackage{xr}
\usepackage{bbm}

\DeclareMathOperator*{\argmax}{arg\,max}

\newcommand{\citet}[1]{\citeauthor{#1} \shortcite{#1}}
\newcommand{\citep}{\cite}
\newcommand{\randeps}{\boldsymbol{\epsilon}}

\title{Ensembles of Locally Independent Prediction Models}
\author{
  Andrew Slavin Ross,\textsuperscript{\rm 1} Weiwei Pan,\textsuperscript{\rm 1} Leo Anthony Celi,\textsuperscript{\rm 2} Finale Doshi-Velez\textsuperscript{\rm 1} \\
\textsuperscript{\rm 1}Paulson School of Engineering and Applied Sciences, Harvard University, Cambridge, MA 02138, USA\\
\textsuperscript{\rm 2}Massachusetts Institute of Technology, Cambridge, MA, 02139, USA\\
  andrew\_ross@g.harvard.edu, weiweipan@g.harvard.edu, lceli@mit.edu, finale@seas.harvard.edu
}

\begin{document}

\maketitle

\begin{abstract}
  Ensembles depend on diversity for improved performance.
  Many ensemble training methods, therefore, attempt to optimize for diversity,
  which they almost always define in terms of differences in training set predictions.
  In this paper, however, we demonstrate the diversity of predictions on the training set
  does not necessarily imply diversity under mild covariate shift,
  which can harm generalization in practical settings.
  To address this issue, we introduce a new diversity metric and associated
  method of training ensembles of models that extrapolate differently on local
  patches of the data manifold. Across a variety of synthetic and real-world
  tasks, we find that our method improves generalization and diversity in
  qualitatively novel ways, especially under data limits and covariate shift.
\end{abstract}

\section{Introduction}

An ensemble is generally more accurate than its constituent models.
However, for this to hold true, those models must make
different errors on unseen data \citep{hansen1990neural,dietterich2000ensemble}.
This is often described as the ensemble's ``diversity.''

Despite diversity's well-recognized importance, there is no firm consensus
on how best to foster it.  Some procedures encourage it
implicitly, e.g. by training models with different inputs
\citep{breiman1996bagging}, while others explicitly optimize for proxies
\citep{liu1999simultaneous} that tend to be functions of differences in
training set predictions \citep{kuncheva2003measures,brown2005diversity}.

However, there has been increasing criticism of supervised machine learning for
focusing too exclusively on cases where training and testing data are drawn
from the same distribution \citep{percytalk}. In many real-world settings, this
assumption does not hold, e.g. due to natural covariate shift over time
\citep{sugiyama2017dataset} or selection bias in data collection
\citep{zadrozny2004learning}.  Intuitively, we might hope that a ``diverse''
ensemble would more easily adapt to such problems, since ideally different
members would be robust to different shifts.  In this paper, however, we find
that diverse ensemble methods that only encourage differences in training
predictions often perform poorly under mild drift between training and test, in
large part because models are not incentivized to make different predictions
where there is no data.  We also find that ensemble methods that directly
optimize for diverse training predictions face inherent tradeoffs between
diversity and accuracy and can be very sensitive to hyperparameters.

To resolve these issues, we make two main contributions, specifically (1) a
novel and differentiable diversity measure, defined as a formal proxy for the
ability of classifiers to \emph{extrapolate} differently away from data, and
(2) a method for training an ensemble of classifiers to be diverse by this
measure, which we hypothesize will lead to more robust predictions under
distributional shifts with no inherent tradeoffs between diversity and accuracy
except those imposed by the dataset. We find this hypothesis holds on a range
of synthetic and real-world prediction tasks.

\section{Related Work}

Ensembling is a well-established subfield of supervised learning
\citep{breiman1996bagging,breiman2001random,ho1995random,schapire1990strength},
and one of its important lessons is that model diversity is a necessary
condition for creating predictive and robust ensembles~\citep{krogh1995neural}.
There are a number of methods for fostering diversity,
which can be roughly divided into two categories: those that implicitly promote diversity by
random modifications to training conditions, and those that explicitly
promote it by deliberate modifications to the objective function.

Some implicit diversity methods operate by introducing stochasticity into
which models see which parts of the data, e.g. by randomly
resampling training examples~\citep{breiman1996bagging} or subsets of input
features~\citep{breiman2001random}.  Others exploit model
parameter stochasticity, e.g. by retraining from different
initializations~\citep{kolen1991back} or
sampling from parameter snapshots saved during individual training
cycles~\citep{huang2017snapshot}.

Methods that explicitly encourage diversity include boosting
\citep{schapire1990strength,freund1997decision}, which sequentially modifies
the objective function of each model to specialize on previous models'
mistakes, or methods like negative correlation learning~\citep{liu1999simultaneous}
amended cross-entropy~\citep{shoham2019amended}, and DPPs over non-maximal
predictions~\cite{pang2019improving}, which simulateously train models with
penalities on both individual errors and pairwise similarities. Finally,
methods such as Diverse Ensemble Evolution~\citep{NIPS2018_7831} and
Competition of Experts~\citep{parascandolo2017learning} use explicit techniques
to encourage models to specialize in different regions of input space.

Although at first glance these diverse training techniques seem
quite diverse themselves, they are all similar in a crucial respect: they
encourage diversity in terms of training set predictions.  In the machine
learning fairness, adversarial robustness, and explainability communities,
however, there has been increasing movement away from the assumption that train is similar to test. For
example, many methods for locally explaining ML predictions literally present
simplified approximations of how models extrapolate away from given points
\cite{baehrens2010explain,ribeiro2016should,ross2017right}, while adversarial
attacks (and defenses) exploit (and mitigate) pathological extrapolation behavior
\cite{szegedy2013intriguing,madry2017towards}, sometimes in an ensemble setting
\cite{tramer2017ensemble}.  Although our focus is not explicitly on
explanability or adversarial robustness, our method can be seen as
a reapplication of techniques in those subfields to the problem of ensemble
diversity.

Also related is the subfield of streaming data, which sometimes uses ensemble
diversity metrics as a criteria for deciding when covariates have shifted
sufficiently to warrant retraining
\citep{brzezinski2016ensemble,krawczyk2017ensemble}. Although our focus remains
on non-streaming classification, the method we introduce may be applicable to
that domain.

\section{Method}

In this section, building on \citet{ross2018learning}, we define our diversity measure and training
procedure, beginning with notation.  We use $x$ to denote $D$-dimensional
inputs, which are supported over an input space $\Omega_x \subseteq
\mathbb{R}^D$. We use $y$ to denote prediction targets in an output space
$\Omega_y$. In this paper, $\Omega_y$ will be $\mathbb{R}$, and we focus on the
case where it represents a log-odds used for binary classification, but our
method can be generalized to classification or regression in $\mathbb{R}^K$
given any notion of distance between outputs.  We seek to learn prediction
models $f(\cdot;\theta): \Omega_x \to \Omega_y$ (parameterized by $\theta$)
that estimate $y$ from $x$.  We assume these models $f$ are differentiable with
respect to $x$ and $\theta$ (which is true for linear models and neural
networks).

In addition, we suppose a joint distribution over inputs and targets $p(x,y)$
and a distribution $p(y|f(x;\theta))$ quantifying the likelihood of the
observed target given the model prediction. Typically, during training, we
seek model parameters that maximize the likelihood of the observed data,
$\mathbb{E}_{p(x,y)} \left[ \log p(y|f(x;\theta)) \right]$.

\subsection{Diversity Measure: Local Independence}\label{sec:diverse}

We now introduce a model diversity measure
that quantifies how differently two models generalize over small patches of the
data manifold $\Omega_x$.
Formally, we define an $\epsilon$-neighborhood of $x$, denoted
$N_\epsilon(x)$, on the data manifold to be the intersection of an
$\epsilon$-ball centered at $x$ in the input space, $\mathcal{B}_\epsilon(x)
\subset \mathbb{R}^D$, and the data manifold: $N_\epsilon(x) =
\mathcal{B}_\epsilon(x)\cap \Omega$. We capture the notion of generalization
difference on a small neighborhood of $x$ through an
intuitive geometric condition: we say that two functions $f$ and $g$
generalize maximally differently at $x$ if $f$ is invariant in the direction
of of the greatest change in $g$ (or vice versa) within an
$\epsilon$-neighborhood around $x$. That is:
\begin{align}
f\left(x\right) = f\left(x_{g_{\text{max}}}\right), \quad \text{for all } \epsilon' < \epsilon, \label{eqn:invt}
\end{align}
where we define $x_{g_{\text{max}}} = \displaystyle\argmax_{x' \in
N_{\epsilon'}(x)} g(x')$.  In other words, perturbing $x$ by small amounts to
increase $g$ inside $N_\epsilon$ does not change the value of $f$. In the case
that a choice of $\epsilon$ exists to satisfy Equation \ref{eqn:invt}, we say
that $f$ is \emph{locally independent at $x$}.  We call $f$ and $g$
\emph{locally independent} without qualification if for every $x\in\Omega_x$
the functions $f$ and $g$ are locally independent at $x$ for some choice of
$\epsilon$. We note that in order for the right-hand side expression of
\ref{eqn:invt} to be well-defined, we assume that the gradient of $g$ is not
zero at $x$ and that $\epsilon$ is chosen to be small enough that $g$ is convex
or concave over $N_{\epsilon}(x)$.

In the case that $f$ and $g$ are classifiers, local independence intuitively
implies a kind of dissimilarity between their decision boundaries. For example,
if $f$ and $g$ are linear and the data manifold is Euclidean, then $f$ and $g$
are locally independent if and only if their decision boundaries are
orthogonal.

This definition motivates the formulation of a diversity measure,
$\mathtt{IndepErr}(f, g)$, quantifying how far $f$ and $g$ are from being
locally independent: 
\begin{equation}\label{eq:adverror} \footnotesize{
  \mathtt{IndepErr}(f, g) \equiv \mathbb{E}\left[\left(f\left(x_{g_{\text{max}}}\right) - f\left(x\right)\right)^2\right].
} \end{equation}

\subsection{Local Independence Training (LIT)}\label{sec:diverse}

Using Equation~\ref{eq:adverror}, we can formulate an ensemble-wide loss
function $\mathcal{L}$ for a set of models $\{\theta_m\}$ as follows, which we
call local independence training:
\begin{equation} \label{eqn:learning_obj} \footnotesize{
\begin{split}
\mathcal{L}(\{\theta_m\}) = &\sum_{m} \mathbb{E}_{p(x,y)} \left[ -\log p(y|f(x;\theta_m)) \right] \\
&+ \lambda \sum_{\ell \neq m} \mathtt{IndepErr}(f(\cdot; \theta_m), f(\cdot; \theta_{\ell})).
\end{split}
} \end{equation}
The first term encourages each model $f_m$ to be predictive and the second
encourages diversity in terms of $\mathtt{IndepErr}$ (with a strength hyperparameter $\lambda$).
Computing $\mathtt{IndepErr}$ exactly, however, is challenging, because it requires an inner
optimization of $g$.  Although it can be closely approximated for fixed small
$\epsilon$ with projected gradient descent as in adversarial training \citep{madry2017towards}, that
procedure is computationally intensive.  If we let $\epsilon \to 0$, however,
we can approximate $x_{g_{\text{max}}}$ by a fairly simple equation that only
needs to compute $\nabla g$ once per $x$. In particular, we observe
that under certain smoothness assumptions on $g$, with
unconstrained $\Omega_x$,\footnote{The simplifying assumption that $N_{\epsilon}(x) \approx \mathcal{B}_{\epsilon}(x)$ in a local neighborhood around $x$ is significant, though not always inappropriate. We discuss both limitations and generalizations in Section~\ref{SUPP-sec:manifolds}.} and as $\epsilon \to 0$, we can make the approximation
\begin{equation}\label{eq:gradpush}
    x_{g_{\text{max}}} \approx x + \epsilon \nabla g(x).
\end{equation}
Assuming similar smoothness assumptions on $f$ (so we can replace it by its
first-order Taylor expansion), we see that
\begin{equation} \footnotesize{
\begin{split}\label{eq:gradderivation}
  f(x_{g_{\text{max}}}&) - f(x)\\
    &\approx f(x + \epsilon \nabla g(x)) - f(x) \\
    &= \left[f(x) + \epsilon \nabla f(x) \cdot \nabla g(x) + \mathcal{O}(\epsilon^2) \right]- f(x) \\
    &\approx \epsilon \nabla f(x) \cdot \nabla g(x).
\end{split}
} \end{equation}
In other words, the independence error between $f$ and $g$ is approximately
equal to the dot product of their gradients $\nabla f(x) \cdot \nabla
g(x)$. Empirically, we find it helpful to normalize the dot product and work in
terms of cosine similarity $\cos(\nabla f(x), \nabla
g(x)) \equiv \frac{\nabla f(x) \cdot \nabla g(x)}{||\nabla f(x)||_2 ||\nabla
g(x)||_2} \in [-1,1]$. We also add a
small constant value to the denominator to prevent numerical underflow. 

\paragraph{Alternate statistical formulation:} As another way of obtaining this cosine similarity approximation,
suppose we sample small perturbations $\randeps{\sim}
\mathcal{N}(0,\sigma^2 \mathbbm{1})$ and evaluate $f(x+\randeps){-}f(x)$
and $g(x+\boldsymbol{\epsilon}){-}g(x)$. As $\sigma{\to}0$, these differences approach $\randeps \cdot \nabla
f(x)$ and $\boldsymbol{\epsilon} \cdot \nabla g(x)$, which are 1D Gaussian
random variables whose correlation is given by $\cos(\nabla f(x), \nabla g(x))$ and whose mutual information is
${-}\textstyle{\frac{1}{2}}\ln(1-\cos^2(\nabla f(x), \nabla g(x)))$ per Section \ref{SUPP-sec:loc-indep}. Therefore,
making the input gradients of $f$ and $g$ orthogonal is equivalent to enforcing
statistical independence between their outputs when we perturb $x$ with samples
from $\mathcal{N}(0,\sigma^2 \mathbbm{1})$ as $\sigma\to0$. This could be used
as an alternate definition of ``local independence.''

\paragraph{Final LIT objective term:} Motivated by the approximations and discussion above, we substitute
\begin{equation}\label{eq:cosgrad} \footnotesize{
\begin{split}
  \mathtt{CosIndepErr}(f,g) \equiv \mathbb{E} \left[\cos^2(\nabla f(x),
  \nabla g(x))\right] \\
\end{split}
}\end{equation}
into our ensemble loss from Equation~\eqref{eqn:learning_obj}, which gives us a final loss function
\begin{equation}\label{eq:final-obj}\footnotesize{
\begin{split}
\mathcal{L}(\{\theta_m\}) = &\sum_{m} \mathbb{E}_{p(x,y)} \left[ -\log p(y|f(x;\theta_m)) \right] \\
  &+ \lambda \sum_{\ell \neq m} \mathbb{E}_{p(x)}\left[\cos^2(\nabla f(x; \theta_m), \nabla f(x; \theta_{\ell}))\right].
\end{split}
}\end{equation}
Note that we will sometimes abbreviate $\mathtt{CosIndepErr}$ as
$\nabla_{\cos^2}$.  In Section~\ref{sec:synth} as well as Figure
\ref{SUPP-fig:rho-vs-cos}, we show that
$\mathtt{CosIndepErr}$ is meaningfully correlated with other diversity measures
and therefore may be useful in its own right, independently of its use within a
loss function.

\section{Experiments} \label{sec:synth}

On synthetic data, we show that ensembles trained with LIT exhibit more
diversity in extrapolation behavior. On a range of benchmark datasets, we show
that the extrapolation diversity in LIT ensembles corresponds to improved
predictive performance on test data that are distributed differently than train
data. Finally, in a medical data case study, we show that models in LIT
ensembles correspond to qualitatively different and clinically meaningful
explanations of the data.

\paragraph{Training:} For the experiments that follow, we use 256-unit single hidden layer fully
connected neural networks with rectifier activations, trained in Tensorflow with Adam.
For the real-data experiments, we use dropout and L2 weight decay with a penalty of 0.0001.
Code to replicate all experiments is available at \texttt{https://github.com/dtak/lit}.

\paragraph{Baselines:} We test local independence training (``LIT'') against random
restarts (``RRs''), bagging \citep{breiman1996bagging} (``Bag''), AdaBoost
\citep{hastie2009multi} (``Ada''), 0-1 squared loss negative correlation learning
\citet{liu1999simultaneous} (``NCL''), and amended cross-entropy
\citep{shoham2019amended} (``ACE'').  We omit \citet{NIPS2018_7831} and
\citet{parascandolo2017learning} which require more complex inner submodular or
adversarial optimization steps, but note that because they also operationalize
diversity as making different errors on training points, we expect the results to be
qualitatively similar to ACE and NCL.

\paragraph{Hyperparameters:} For our non-synthetic results, we test all methods with
ensemble sizes in $\{2, 3, 5, 8, 13\}$, and all methods with regularization
parameters $\lambda$ (LIT, ACE, and NCL) with 16 logarithmically spaced values
between $10^{-4}$ and $10^1$, using validation AUC to select the best
performing model (except when examining how results vary with $\lambda$ or
size). For each hyperparameter setting and method, we run 10 full random
restarts (though within each restart, different methods are tested against the
same split), and present mean results with standard deviation errorbars.

\subsection{Conceptual Demonstration}

\begin{figure}[htb!]
  \centering
  \includegraphics[width=\columnwidth]{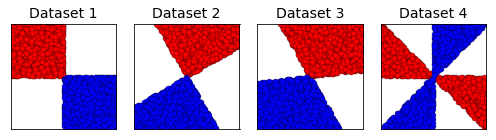}
  \caption{2D synthetic datasets with gaps. We argue that ``diverse'' ensemble methods applied to these datasets should produce accurate models with different decision boundaries.}
  \label{fig:2d-toy-examples}
\end{figure}

To provide an initial demonstration of our method and the limitations of
training set prediction diversity, we present several sets of 2D synthetic
examples in Figure \ref{fig:2d-toy-examples}.  These 2D examples are
constructed to have data distributions that satisfy our assumption that
$N_\epsilon(x) \approx \mathcal{B}_\epsilon(x)$ locally around almost all of
the points, but nevertheless contain significant gaps.  These gaps result in the
possibility of learning multiple classifiers that have perfect accuracy on the
training set but behave differently when extrapolating. Indeed, in all of these
examples, if we have just two classifiers, they can completely agree on
training and completely disagree in the extrapolation regions.

\begin{figure}[htb!]
  \centering
  \includegraphics[width=\columnwidth]{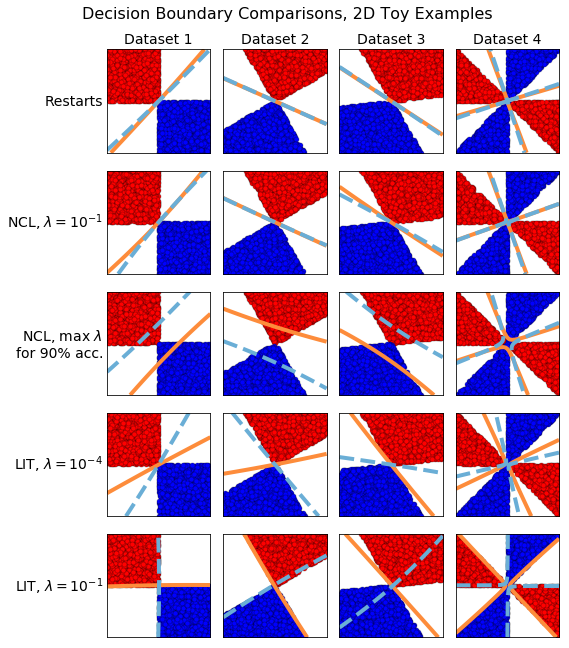}
  \caption{Comparison of local independence training, random restarts and NCL
  on toy 2D datasets. For each ensemble, the first model's decision boundary
  is plotted in orange and the other in dashed blue. Both NCL and LIT are capable
  of producing variation, but in qualitatively different ways.}
  \label{fig:2d-toy-results}
\end{figure}

\begin{table*}[htb!]
\centering
\small
\resizebox{2.1\columnwidth}{!}{
\begin{tabular}{|c||c|c|c||c|c|c||c|c|c||c|c|c||c|c|c|}
\multicolumn{16}{c}{Random Split} \\ \hline
\multirow{2}{*}{Method} & \multicolumn{3}{c||}{Mushroom} & \multicolumn{3}{c||}{Ionosphere} & \multicolumn{3}{c||}{Sonar} & \multicolumn{3}{c||}{SPECTF} & \multicolumn{3}{c|}{Electricity} \\ \cline{2-16}
 & AUC & $\rho_{av}$ & $\nabla_{\cos^2}$ & AUC & $\rho_{av}$ & $\nabla_{\cos^2}$ & AUC & $\rho_{av}$ & $\nabla_{\cos^2}$ & AUC & $\rho_{av}$ & $\nabla_{\cos^2}$ & AUC & $\rho_{av}$ & $\nabla_{\cos^2}$ \\ \hline
RRs & \textbf{1.0} & $\text{1}{\pm}\text{.1}$ & $\text{.9}{\pm}\text{0}$ & $\text{.95}{\pm}\text{.03}$ & $\text{.9}{\pm}\text{.1}$ & $\text{1}{\pm}\text{0}$ & $\text{.91}{\pm}\text{.06}$ & $\text{.9}{\pm}\text{.1}$ & $\text{1}{\pm}\text{0}$ & $\textbf{.80}{\pm}\text{.06}$ & $\text{.9}{\pm}\text{.1}$ & $\text{1}{\pm}\text{0}$ & $\text{.87}{\pm}\text{.00}$ & $\text{1}{\pm}\text{0}$ & $\text{1}{\pm}\text{0}$\\ \hline
Bag & \textbf{1.0} & $\text{1}{\pm}\text{0}$ & $\text{.9}{\pm}\text{0}$ & $\text{.96}{\pm}\text{.02}$ & $\text{.7}{\pm}\text{.1}$ & $\text{.5}{\pm}\text{.1}$ & $\text{.90}{\pm}\text{.06}$ & $\text{.5}{\pm}\text{.2}$ & $\text{.5}{\pm}\text{.1}$ & $\textbf{.80}{\pm}\text{.05}$ & $\text{.6}{\pm}\text{.1}$ & $\text{.4}{\pm}\text{.1}$ & $\text{.87}{\pm}\text{.00}$ & $\text{.9}{\pm}\text{0}$ & $\text{1}{\pm}\text{0}$\\ \hline
Ada & \textbf{1.0} & --- & --- & $\text{.95}{\pm}\text{.03}$ & --- & --- & $\text{.91}{\pm}\text{.06}$ & --- & --- & $\textbf{.80}{\pm}\text{.06}$ & --- & --- & $\textbf{.88}{\pm}\text{.00}$ & $\text{.2}{\pm}\text{0}$ & $\text{.2}{\pm}\text{.1}$\\ \hline
NCL & \textbf{1.0} & $\text{1}{\pm}\text{0}$ & $\text{.8}{\pm}\text{0}$ & $\text{.96}{\pm}\text{.04}$ & $\text{.6}{\pm}\text{.5}$ & $\text{.7}{\pm}\text{.3}$ & $\textbf{.91}{\pm}\text{.06}$ & $\text{.6}{\pm}\text{.5}$ & $\text{.7}{\pm}\text{.4}$ & $\textbf{.80}{\pm}\text{.07}$ & $\text{.6}{\pm}\text{.5}$ & $\text{.7}{\pm}\text{.3}$ & $\text{.87}{\pm}\text{.00}$ & $\text{.4}{\pm}\text{.1}$ & $\text{.6}{\pm}\text{0}$\\ \hline
ACE & \textbf{1.0} & $\text{1}{\pm}\text{0}$ & $\text{.9}{\pm}\text{0}$ & $\text{.94}{\pm}\text{.04}$ & $\text{.8}{\pm}\text{.3}$ & $\text{.9}{\pm}\text{.2}$ & $\text{.90}{\pm}\text{.06}$ & $\text{.9}{\pm}\text{.2}$ & $\text{1}{\pm}\text{.1}$ & $\textbf{.79}{\pm}\text{.06}$ & $\text{.8}{\pm}\text{.4}$ & $\text{.9}{\pm}\text{.2}$ & $\text{.87}{\pm}\text{.00}$ & $\text{.9}{\pm}\text{0}$ & $\text{1}{\pm}\text{0}$\\ \hline
LIT & \textbf{1.0} & $\text{.9}{\pm}\text{.1}$ & $\text{0}{\pm}\text{0}$ & $\textbf{.98}{\pm}\text{.01}$ & $\text{.3}{\pm}\text{.1}$ & $\text{0}{\pm}\text{0}$ & $\textbf{.92}{\pm}\text{.05}$ & $\text{.5}{\pm}\text{.2}$ & $\text{0}{\pm}\text{0}$ & $\textbf{.81}{\pm}\text{.06}$ & $\text{.4}{\pm}\text{.1}$ & $\text{0}{\pm}\text{0}$ & $\text{.87}{\pm}\text{.00}$ & $\text{.9}{\pm}\text{0}$ & $\text{.3}{\pm}\text{.1}$\\ \hline
\multicolumn{16}{c}{Extrapolation Split} \\ \hline
\multirow{2}{*}{Method} & \multicolumn{3}{c||}{Mushroom} & \multicolumn{3}{c||}{Ionosphere} & \multicolumn{3}{c||}{Sonar} & \multicolumn{3}{c||}{SPECTF} & \multicolumn{3}{c|}{Electricity} \\ \cline{2-16}
 & AUC & $\rho_{av}$ & $\nabla_{\cos^2}$ & AUC & $\rho_{av}$ & $\nabla_{\cos^2}$ & AUC & $\rho_{av}$ & $\nabla_{\cos^2}$ & AUC & $\rho_{av}$ & $\nabla_{\cos^2}$ & AUC & $\rho_{av}$ & $\nabla_{\cos^2}$ \\ \hline
RRs & $\text{.92}{\pm}\text{.00}$ & $\text{.9}{\pm}\text{0}$ & $\text{.8}{\pm}\text{0}$ & $\text{.87}{\pm}\text{.02}$ & $\text{1}{\pm}\text{0}$ & $\text{1}{\pm}\text{0}$ & $\text{.81}{\pm}\text{.02}$ & $\text{1}{\pm}\text{0}$ & $\text{1}{\pm}\text{0}$ & $\textbf{.83}{\pm}\text{.05}$ & $\text{1}{\pm}\text{0}$ & $\text{1}{\pm}\text{0}$ & $\text{.86}{\pm}\text{.00}$ & $\text{1}{\pm}\text{0}$ & $\text{1}{\pm}\text{0}$\\ \hline
Bag & $\text{.91}{\pm}\text{.00}$ & $\text{.9}{\pm}\text{0}$ & $\text{.9}{\pm}\text{0}$ & $\text{.89}{\pm}\text{.04}$ & $\text{.6}{\pm}\text{.1}$ & $\text{.5}{\pm}\text{.1}$ & $\textbf{.82}{\pm}\text{.03}$ & $\text{.7}{\pm}\text{.1}$ & $\text{.6}{\pm}\text{0}$ & $\textbf{.83}{\pm}\text{.05}$ & $\text{.6}{\pm}\text{.1}$ & $\text{.4}{\pm}\text{0}$ & $\text{.86}{\pm}\text{.00}$ & $\text{.9}{\pm}\text{0}$ & $\text{.9}{\pm}\text{0}$\\ \hline
Ada & $\text{.92}{\pm}\text{.01}$ & --- & --- & $\text{.87}{\pm}\text{.02}$ & --- & --- & $\text{.81}{\pm}\text{.03}$ & --- & --- & $\textbf{.83}{\pm}\text{.05}$ & --- & --- & $\text{.86}{\pm}\text{.00}$ & $\text{.3}{\pm}\text{.1}$ & $\text{.3}{\pm}\text{.2}$\\ \hline
NCL & $\text{.94}{\pm}\text{.01}$ & $\text{.6}{\pm}\text{.2}$ & $\text{.6}{\pm}\text{.1}$ & $\text{.90}{\pm}\text{.02}$ & $\text{.8}{\pm}\text{.3}$ & $\text{.9}{\pm}\text{.2}$ & $\text{.78}{\pm}\text{.06}$ & $\text{.5}{\pm}\text{.5}$ & $\text{.6}{\pm}\text{.3}$ & $\text{.81}{\pm}\text{.12}$ & $\text{.5}{\pm}\text{.6}$ & $\text{.7}{\pm}\text{.3}$ & $\text{.86}{\pm}\text{.00}$ & $\text{.9}{\pm}\text{.2}$ & $\text{1}{\pm}\text{.1}$\\ \hline
ACE & $\text{.92}{\pm}\text{.00}$ & $\text{.9}{\pm}\text{0}$ & $\text{.8}{\pm}\text{0}$ & $\text{.90}{\pm}\text{.03}$ & $\text{.3}{\pm}\text{.4}$ & $\text{.5}{\pm}\text{.3}$ & $\text{.77}{\pm}\text{.06}$ & $\text{.6}{\pm}\text{.5}$ & $\text{.7}{\pm}\text{.3}$ & $\text{.72}{\pm}\text{.16}$ & $\text{.5}{\pm}\text{.6}$ & $\text{.7}{\pm}\text{.4}$ & $\text{.86}{\pm}\text{.00}$ & $\text{1}{\pm}\text{0}$ & $\text{1}{\pm}\text{0}$\\ \hline
LIT & $\textbf{.96}{\pm}\text{.01}$ & $\text{.3}{\pm}\text{.1}$ & $\text{0}{\pm}\text{0}$ & $\textbf{.96}{\pm}\text{.02}$ & $\text{.2}{\pm}\text{.1}$ & $\text{0}{\pm}\text{0}$ & $\textbf{.81}{\pm}\text{.03}$ & $\text{.5}{\pm}\text{.1}$ & $\text{0}{\pm}\text{0}$ & $\textbf{.84}{\pm}\text{.05}$ & $\text{.4}{\pm}\text{.1}$ & $\text{0}{\pm}\text{0}$ & $\textbf{.87}{\pm}\text{.00}$ & $\text{.4}{\pm}\text{.2}$ & $\text{0}{\pm}\text{0}$\\ \hline
\end{tabular}
  }
  \caption{Benchmark classification results in both the normal prediction task (top) and the extrapolation task (bottom) over 10 reruns, with errorbars based on standard deviations and bolding based on standard error overlap. On random splits, LIT offers modest AUC improvements over random restarts, on par with other ensemble methods. On extrapolation splits, however, LIT tends to achieve higher AUC. In both cases, LIT almost always exhibits low pairwise Pearson correlation between heldout model errors ($\rho_{av}$), and for other methods, $\rho_{av}$ roughly matches pairwise gradient cosine similarity ($\nabla_{\cos^2}$).
  }
  \label{table:uci-auc}
\end{table*}

In Figure \ref{fig:2d-toy-results}, we compare the neural network decision
boundaries learned by random restarts, local independence training, and
negative correlation learning (NCL) on these examples (we use NCL as a
state-of-the-art example of an approach that defines diversity with respect to
training predictions).  Starting with the top and bottom two rows (random
restarts and LIT), we find that random restarts give us essentially identical
models, whereas LIT outputs models with meaningfully different decision
boundaries even at values of $\lambda$ that are very low compared to its
prediction loss term.  This is in large part because on most of these tasks
(except Dataset 3), there is very little tradeoff to learning a near-orthogonal
boundary. At larger $\lambda$, LIT outputs decision boundaries that are
completely orthogonal (at the cost of a slight accuracy reduction on Dataset
3).

NCL had more complicated behavior, in large part because of its built-in
tradeoff between accuracy and diversity. At low values of $\lambda$
(second from top), we found that NCL produced models with identical
decision boundaries, suggesting that training ignored the
diversity term. At $\lambda \geq 2$, the
predictive performance of one model fell to random guessing,
suggesting that training ignored the accuracy term. So in order
to obtain meaningfully diverse but accurate NCL models, we iteratively searched for the highest
value of $\lambda$ at which NCL would still return two models at least
90\% accurate on the training set (by exponentially shrinking a window between $\lambda=1$
and $\lambda=2$ for 10 iterations). What we found (middle row) is that
NCL learned to translate its decision boundaries within the support of the training data (incurring an
initially modest accuracy cost due to the geometry of the problem) but not
modify them outside the training support.
Although this kind of diversity is not
necessarily bad (since the \emph{ensemble} accuracy remains perfect), it is
qualitatively different from the kind of diversity encouraged by LIT---and only
emerges at carefully chosen hyperparameter values. The main takeaway from this
set of synthetic examples is that methods that encourage diverse extrapolation
(like LIT) can produce significantly different ensembles than methods that
encourage diverse prediction (like NCL).

\subsection{Classification Benchmarks}

Next, we test our method on several standard binary classification datasets
from the UCI and MOA repositories \citep{uci,moa}.  These are
\texttt{mushroom}, \texttt{ionosphere}, \texttt{sonar}, \texttt{spectf}, and
\texttt{electricity} (with categorical features one-hot encoded, and all
features z-scored).  For all datasets, we randomly select 80\% of the dataset
for training and 20\% for test, then take an additional 20\% split of the
training set to use for validation.  In addition to random splits, we also
introduce an \emph{extrapolation} task, where instead of splitting datasets
randomly, we train on the 50\% of points closest to the origin (i.e. where
$||x||_2$ is less than its median value) and validate/test on the remaining
points (which are furthest from the origin).  This test is meant to evaluate
robustness to covariate shift.

For each ensemble, we measure heldout AUC and accuracy, our diversity metric
\texttt{CosIndepErr} (abbreviated as $\nabla_{\cos^2}$), and several classic
diversity metrics ($\rho_{av}$, $Q_{av}$, and $\kappa$) defined by
\citet{kuncheva2003measures}.  Table \ref{table:uci-auc} compares heldout AUC,
$\rho_{av}$, and $\nabla_{\cos^2}$ after cross-validating $\lambda$ and the
ensemble size. More complete enumerations of AUC, accuracy, and diversity
metrics are shown in Figures \ref{SUPP-fig:all-aucs} and
\ref{SUPP-fig:rho-vs-cos}. In general, we find that LIT is competitive on
random splits, strongest on extrapolation, and significantly improves heldout
prediction diversity across the board. We also find that $\nabla_{\cos^2}$ is
meaningfully related to other diversity metrics for all models that do not
optimize for it.

\subsection{ICU Mortality Case Study} \label{sec:clinical}

As a final set of experiments, we run a more in-depth case study on a real world
clinical application.  In particular, we predict in-hospital mortality for a
cohort of $n=1,053,490$ patient visits extracted from the MIMIC-III database
\citep{mimic3} based on on labs, vital signs, and basic demographics. We follow
the same cohort selection and feature selection process as
\citet{ghassemi2017predicting}.
In addition to this full cohort, we also test on a limited data task where we
restrict the size of the training set to $n=1000$ to measure robustness.

\begin{figure}[htb!]
  \centering
  \includegraphics[width=\columnwidth]{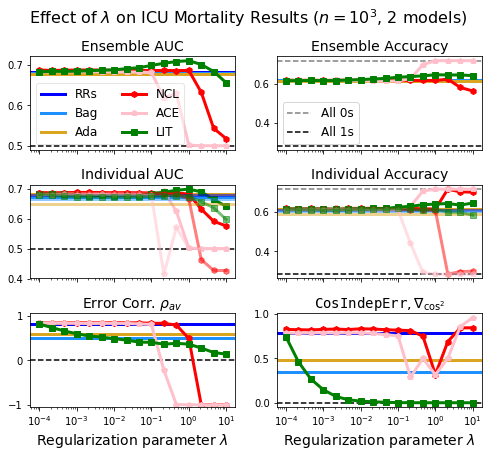}
  \caption{Changes in individual AUC/accuracy and ensemble diversity with
  $\lambda$ for two-model ensembles on the ICU mortality dataset (averaged
  across 10 reruns, error-bars omitted for clarity).  For NCL and ACE, there is
  a wide low-$\lambda$ regime where they are indistinguishable from random
  restarts. This is followed by a very brief window of meaningful diversity
  (around $\lambda=1$ for NCL, slightly lower for ACE), after which both
  methods output pairs of models which always predict 0 and 1 (respectively),
  as shown by the error correlation dropping to -1. LIT, on the other hand,
  exhibits smooth drops in individual model predictive performance, with error
  correlation falling towards 0. Results for other ensemble sizes were
  qualitatively similar.}
  \label{fig:icu-2models}
\end{figure}

\begin{figure}[htb!]
  \centering
  \includegraphics[width=0.9\columnwidth]{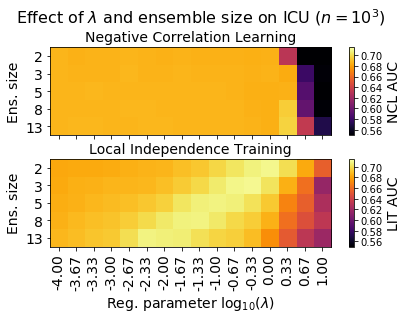}
  \caption{Another exploration of the effect of ensemble size and $\lambda$ on
  ICU mortality predictions.  In particular, we find that for LIT on this
  dataset, the optimal value of $\lambda$ depends on the ensemble size in a
  roughly log-linear relationship.  Because $D$-dimensional datasets can
  support a maximum of $D$ locally independent models (and only one model if
  the data completely determines the decision boundary), it is intuitive that
  there should be an optimal value.  For NCL, we also observe an optimal value
  near $10^{0.33}$, but with a less clear relationship to ensemble size and
  very steep dropoff to random guessing at slightly higher $\lambda$.}
  \label{fig:icu-hypers}
\end{figure}

We visualize the results of these experiments in several ways to help
tease out the effects of $\lambda$, ensemble size, and dataset size on
individual and ensemble predictive performance, diversity, and model
explanations. Table \ref{tbl:icu-aucs} shows overall performance and diversity
metrics for these two tasks after cross-validation, along with the most common
values of $\lambda$ and ensemble size selected for each method. Drilling into
the $n=1000$ results, Figure \ref{fig:icu-2models} visualizes how multiple
metrics for performance (AUC and accuracy) and diversity ($\rho_{av}$ and
$\nabla_{\cos^2}$) change with $\lambda$, while Figure \ref{fig:icu-hypers}
visualizes the relationship between optimal $\lambda$ and ensemble size.

\begin{figure}[htb!]
  \centering
  \includegraphics[width=\columnwidth]{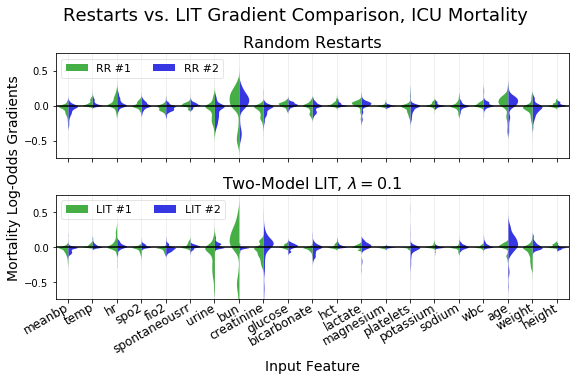}
  \caption{Differences in cross-patient gradient distributions of ICU mortality
  prediction models for random restart and locally independent ensembles (similar plots for other methods are shown in Figure~\ref{SUPP-fig:twomod-grads}).
  Features with mass consistently above the x-axis have positive associations
  with predicted mortality (increasing them increases predicted mortality)
  while those with mass consistently below the x-axis have negative
  associations (decreasing them increases predicted mortality). Distance from
  the x-axis corresponds to the association strength. Models trained normally
  (top) consistently learn positive associations with \texttt{age} and
  \texttt{bun} (blood urea nitrogen; larger values indicate kidney failure) and
  negative associations with \texttt{weight} and \texttt{urine} (low weight is
  correlated with mortality; low urine output also indicates kidney failure or
  internal bleeding). However, they also learn somewhat negative associations with \texttt{creatinine},
  which confused clinicians because high values are another indicator of kidney failure.
  When we trained LIT models, however, we found that \texttt{creatinine} regained its positive association with mortality
  (in model 2), while the other main features were more or less divided up.
  This collinearity between \texttt{creatinine} and \texttt{bun}/\texttt{urine} in
  indicating organ problems (and revealed by LIT) was one of the main insights derived in our
  qualitative evaluation with ICU clinicians.}
  \label{fig:sepsis-gradient-spectra}
\end{figure}

Figure \ref{fig:sepsis-gradient-spectra} (as well as Figures \ref{SUPP-fig:extrap-grads} and \ref{SUPP-fig:twomod-grads})
visualize changes in the marginal distributions of input gradients
for each model in their explanatory sense \cite{baehrens2010explain}. As a
qualitative evaluation, we discussed these explanation differences with two
intensive care unit clinicians and found that LIT revealed meaningful
redundancies in which combinations of features encoded different underlying
conditions.

\begin{table}[htb!]
  \centering
\small
\begin{tabular}{|c||c|c|c|c|c|}
  \multicolumn{6}{c}{ICU Mortality Task, Full Dataset ($n>10^6$)} \\ \hline
Method & AUC & $\rho_{av}$ & $\nabla_{\cos^2}$ & \# & $\lambda$ \\ \hline
RRs & $\text{.750}{\pm}\text{.000}$ & $\text{.9}{\pm}\text{0}$ & $\text{.9}{\pm}\text{0}$ & 13 & ---\\ \hline
Bag & $\text{.751}{\pm}\text{.000}$ & $\text{.9}{\pm}\text{0}$ & $\text{.9}{\pm}\text{0}$ & 8 & ---\\ \hline
Ada & $\textbf{.752}{\pm}\text{.003}$ & $\text{0}{\pm}\text{0}$ & $\text{0}{\pm}\text{0}$ & 8 & ---\\ \hline
ACE & $\text{.750}{\pm}\text{.000}$ & $\text{.9}{\pm}\text{0}$ & $\text{.9}{\pm}\text{0}$ & 13 & $10^{0.33}$\\ \hline
NCL & $\textbf{.753}{\pm}\text{.001}$ & $\text{.3}{\pm}\text{.2}$ & $\text{.2}{\pm}\text{.2}$ & 13 & $10^{0.00}$\\ \hline
LIT & $\text{.750}{\pm}\text{.001}$ & $\text{.8}{\pm}\text{0}$ & $\text{.3}{\pm}\text{0}$ & 3 & $10^{-4.00}$\\ \hline
  \multicolumn{6}{c}{ICU Mortality Task, Limited Slice ($n=10^3$)} \\ \hline
Method & AUC & $\rho_{av}$ & $\nabla_{\cos^2}$ & \# & $\lambda$ \\ \hline
RRs & $\text{.684}{\pm}\text{.001}$ & $\text{.8}{\pm}\text{0}$ & $\text{.8}{\pm}\text{0}$ & 8 & ---\\ \hline
Bag & $\text{.690}{\pm}\text{.002}$ & $\text{.5}{\pm}\text{0}$ & $\text{.3}{\pm}\text{0}$ & 8 & ---\\ \hline
Ada & $\text{.678}{\pm}\text{.003}$ & $\text{.6}{\pm}\text{0}$ & $\text{.5}{\pm}\text{0}$ & 2 & ---\\ \hline
ACE & $\text{.684}{\pm}\text{.001}$ & $\text{.8}{\pm}\text{0}$ & $\text{.8}{\pm}\text{0}$ & 2 & $10^{-2.67}$\\ \hline
NCL & $\text{.697}{\pm}\text{.006}$ & $\text{.2}{\pm}\text{.4}$ & $\text{.6}{\pm}\text{.2}$ & 13 & $10^{0.33}$\\ \hline
LIT & $\textbf{.711}{\pm}\text{.001}$ & $\text{.1}{\pm}\text{0}$ & $\text{0}{\pm}\text{0}$ & 13 & $10^{-2.33}$\\ \hline
\end{tabular}
  \caption{Quantitative results on the ICU mortality prediction task, where
  $\#$ and $\lambda$ signify the most commonly selected values of ensemble size
  and regularization parameter chosen for each method. On the full data task,
  although all methods perform similarly, NCL and AdaBoost edge out slightly,
  and LIT consistently selects its weakest regularization parameter. On the
  limited data task, LIT significantly outperforms baselines, with NCL and
  Bagging in second, ACE indistinguishable from restarts, and significantly
  worse performance for AdaBoost (which overfits).}
  \label{tbl:icu-aucs}
\end{table}

\section{Discussion}\label{sec:discussion}

\paragraph{LIT matches or outperforms other methods, especially under data limits or covariate shift.}
On the UCI datasets under $\text{train}\overset{d}{\approx}\text{test}$
conditions (random splits), LIT always offers at least modest improvements over
random restarts, and often outperforms other baselines.  Under extrapolation
splits, LIT tends to do significantly better. This pattern repeats itself on
the normal vs. data-limited versions of ICU mortality prediction task.  We
hypothesize that on small or selectively restricted datasets, there is
typically more predictive ambiguity, which hurts the generalization of normally
trained ensembles (who consistently make similar guesses on unseen data). LIT
is more robust to these issues.

\paragraph{Gradient cosine similarity can be a meaningful diversity metric.} In
Table~\ref{table:uci-auc} as well as our more complete results in Figure
\ref{SUPP-fig:rho-vs-cos}, we saw that for non-LIT methods, gradient similarity
$\nabla_{\cos^2}$ (which does not require labels to compute) was often similar
in value to error correlation $\rho_{av}$ (as well as the interrater agreement
$\kappa$, or Yule's Q-statistic $Q_{av}$ after a monotonic
transformation---all measures which \emph{do} require labels to compute).  One
potential explanation for this correspondence is that, by our analysis at the
end of Section~\ref{sec:diverse}, $\nabla_{\cos^2}$ can literally be
interpreted as an average squared correlation (between changes in model
predictions over infinitesimal Gaussian perturbations away from each input). We
hypothesize that $\nabla_{\cos^2}$ may be a useful quantity independently of
LIT.

\paragraph{LIT is less sensitive to hyperparameters than baselines, but
ensemble size matters more.} In both our synthetic examples (Figure
\ref{fig:2d-toy-results}) and our ICU mortality results (Figures
\ref{fig:icu-2models} and \ref{fig:icu-hypers}), we found that LIT produced
qualitatively similar (diverse) results over several orders of magnitude of
$\lambda$. NCL, on the other hand, required careful tuning of
$\lambda$ to achieve meaningful diversity (before its performance plummeted).
In line with the results from our synthetic examples, we believe this
difference stems from the fact that NCL's diversity term is formulated as a
direct tradeoff with individual model accuracy, so the balance must be precise,
whereas LIT's diversity term can theoretically be completely independent of individual
model accuracy (which is true by construction in the synthetic examples).  However,
datasets only have the capacity to support a limited number of (mostly or
completely) locally independent models.  On the synthetic datasets, this
capacity was exactly 2, but on real data, it is generally unknown, and it may
be possible to achieve similar results either with a small fully independent
ensemble or a large partially independent ensemble. For example, in Figure
\ref{fig:icu-hypers}, we show that we can achieve similar improvements to ICU
mortality prediction with 2 highly independent ($\lambda=10^0$) models or 13
more weakly independent ($\lambda=10^{-2.33}$) models. We hypothesize that the
trend-line of optimal LIT ensemble size and $\lambda$ may be a useful tool for
characterizing the amount of ambiguity
present in a dataset.

\paragraph{Interpretation of individual LIT models can yield useful dataset
insights.} In Figure \ref{fig:sepsis-gradient-spectra}, we found that in
discussions with ICU clinicians, mortality feature assocations for normally
trained neural networks were somewhat confusing due to hidden collinearities.
LIT models made more clinical sense individually, and the differences between
them helped reveal those collinearities (in particular between elevated levels
of blood urea nitrogen and creatinine). Because LIT ensembles are often optimal
when small, and because individual LIT models are not required to sacrifice
accuracy for diversity, they may enable different and more useful kinds of data
interpretation than other ensemble methods.

\paragraph{Limitations.} LIT does come with restrictions and limitations.  In
particular, we found that it works well for rectifier activations (e.g. ReLU
and softplus\footnote{Although we used ReLU in our quantitative experiments, we
found more consistent behavior in synthetic examples with softplus, perhaps due
to its many-times differentiability.}) but leads to inconsistent behavior with
others (e.g. sigmoid and tanh). This may be related to the linear rather than
saturating extrapolation behavior of rectifiers.  Because it relies on cosine
similarity, LIT is also sensitive to relative changes in feature scaling;
however, in practice this issue can be resolved by standardizing variables
first.

Additionally, our cosine similarity approximation in LIT makes the assumption
that the data manifold is locally similar to $\mathbb{R}^D$ near most inputs.
However, we introduce generalizations in Section~\ref{SUPP-sec:manifolds} to
handle situations where this is not approximately true (such as with image
data).

Finally, LIT requires computing a second derivative (the derivative of the
penalty) during the optimization process, which increases memory usage and
training time; in practice, LIT took approximately 1.5x as long as random
restarts, while NCL took approximately half the time.  However, significant
progress is being made on making higher-order autodifferentiation more
efficient \cite{betancourt2018geometric}, so we can expect improvements.
Also, in cases where LIT achieves high accuracy with a comparatively small
ensemble size (e.g. ICU mortality prediction), overall training time
can remain short if cross-validation terminates early.

\section{Conclusion and Future Work}

In this paper, we presented a novel diversity metric that formalizes the notion
of difference in local extrapolations. Based on this metric we defined an
ensemble method, local independence training, for building ensembles of highly
predictive base models that generalize differently outside the training set. On
datasets we knew supported multiple diverse decision boundaries, we
demonstrated our method's ability to recover them. On real-world datasets with
unknown levels of redundancy, we demonstrated that LIT ensembles perform
competitively on traditional prediction tasks and were more robust to data
scarcity and covariate shift (as measured by training on inliers and testing on
outliers). Finally, in our case study on a clinical prediction task in the
intensive care unit, we provided evidence that the extrapolation diversity
exhibited by LIT ensembles improved data robustness and helped us reach
meaningful clinical insights in conversations with
clinicians.  Together, these results suggest that extrapolation diversity may
be an important quantity for ensemble algorithms to measure and optimize.

There are ample directions for future improvements. For example, it would be
useful to consider methods for aggregating predictions of LIT ensembles using a
more complex mechanism, such as a mixture-of-experts model. Along similar
lines, combining pairwise $\mathtt{IndepErr}$s in more informed way, such as a
determinantal point process penalty~\citep{kulesza2012determinantal} over the
matrix of model similarities, may help us better quantify the diversity of the
ensemble. Another interesting extension of our work would be to prediction
tasks in semi-supervised settings, since labels are generally not required for
computing local independence error. Finally, as we observe in the Section
\ref{sec:discussion}, some datasets seem to support a particular number of
locally independent models. It is worth exploring how to connect this property
to attempts to formally quantify and characterize the complexity or ambiguity
present in a prediction task \citep{lorena2018complex,semenova2019study}.

\section*{Acknowledgements}

WP acknowledges the Harvard Institute for Applied Computational Science for its support. ASR is supported by NIH 1R56MH115187. The authors also wish to thank the anonymous reviewers for helpful feedback.

\bibliographystyle{aaai}
\bibliography{bibliography}

\begin{thebibliography}{}

\bibitem[\protect\citeauthoryear{Baehrens \bgroup et al\mbox.\egroup
  }{2010}]{baehrens2010explain}
Baehrens, D.; Schroeter, T.; Harmeling, S.; Kawanabe, M.; Hansen, K.; and
  M{\"u}ller, K.-R.
\newblock 2010.
\newblock How to explain individual classification decisions.
\newblock {\em Journal of Machine Learning Research} 11(6).

\bibitem[\protect\citeauthoryear{Betancourt}{2018}]{betancourt2018geometric}
Betancourt, M.
\newblock 2018.
\newblock A geometric theory of higher-order automatic differentiation.
\newblock {\em arXiv preprint arXiv:1812.11592}.

\bibitem[\protect\citeauthoryear{Bifet \bgroup et al\mbox.\egroup }{2010}]{moa}
Bifet, A.; Holmes, G.; Kirkby, R.; and Pfahringer, B.
\newblock 2010.
\newblock {MOA:} massive online analysis.
\newblock {\em Journal of Machine Learning Research} 11.

\bibitem[\protect\citeauthoryear{Breiman}{1996}]{breiman1996bagging}
Breiman, L.
\newblock 1996.
\newblock Bagging predictors.
\newblock {\em Machine learning} 24(2).

\bibitem[\protect\citeauthoryear{Breiman}{2001}]{breiman2001random}
Breiman, L.
\newblock 2001.
\newblock Random forests.
\newblock {\em Machine learning} 45(1).

\bibitem[\protect\citeauthoryear{Brown \bgroup et al\mbox.\egroup
  }{2005}]{brown2005diversity}
Brown, G.; Wyatt, J.; Harris, R.; and Yao, X.
\newblock 2005.
\newblock Diversity creation methods: a survey and categorisation.
\newblock {\em Information Fusion}.

\bibitem[\protect\citeauthoryear{Brzezinski and
  Stefanowski}{2016}]{brzezinski2016ensemble}
Brzezinski, D., and Stefanowski, J.
\newblock 2016.
\newblock Ensemble diversity in evolving data streams.
\newblock In {\em International Conference on Discovery Science}.

\bibitem[\protect\citeauthoryear{Dietterich}{2000}]{dietterich2000ensemble}
Dietterich, T.~G.
\newblock 2000.
\newblock Ensemble methods in machine learning.
\newblock In {\em International workshop on multiple classifier systems}.

\bibitem[\protect\citeauthoryear{Freund and
  Schapire}{1997}]{freund1997decision}
Freund, Y., and Schapire, R.~E.
\newblock 1997.
\newblock A decision-theoretic generalization of on-line learning and an
  application to boosting.
\newblock {\em Journal of computer and system sciences} 55(1).

\bibitem[\protect\citeauthoryear{Ghassemi \bgroup et al\mbox.\egroup
  }{2017}]{ghassemi2017predicting}
Ghassemi, M.; Wu, M.; Hughes, M.~C.; Szolovits, P.; and Doshi-Velez, F.
\newblock 2017.
\newblock Predicting intervention onset in the icu with switching state space
  models.
\newblock {\em AMIA Summits on Translational Science Proceedings} 2017.

\bibitem[\protect\citeauthoryear{Hansen and Salamon}{1990}]{hansen1990neural}
Hansen, L.~K., and Salamon, P.
\newblock 1990.
\newblock Neural network ensembles.
\newblock {\em IEEE transactions on pattern analysis and machine intelligence}
  12(10).

\bibitem[\protect\citeauthoryear{Hastie \bgroup et al\mbox.\egroup
  }{2009}]{hastie2009multi}
Hastie, T.; Rosset, S.; Zhu, J.; and Zou, H.
\newblock 2009.
\newblock Multi-class adaboost.
\newblock {\em Statistics and its Interface}.

\bibitem[\protect\citeauthoryear{Ho}{1995}]{ho1995random}
Ho, T.~K.
\newblock 1995.
\newblock Random decision forests.
\newblock In {\em Document analysis and recognition, 1995., proceedings of the
  third international conference on}, volume~1.
\newblock IEEE.

\bibitem[\protect\citeauthoryear{Huang \bgroup et al\mbox.\egroup
  }{2017}]{huang2017snapshot}
Huang, G.; Li, Y.; Pleiss, G.; Liu, Z.; Hopcroft, J.~E.; and Weinberger, K.~Q.
\newblock 2017.
\newblock Snapshot ensembles: Train 1, get m for free.
\newblock {\em arXiv preprint arXiv:1704.00109}.

\bibitem[\protect\citeauthoryear{Johnson \bgroup et al\mbox.\egroup
  }{2016}]{mimic3}
Johnson, A.~E.; Pollard, T.~J.; Shen, L.; Lehman, L.-w.~H.; Feng, M.; Ghassemi,
  M.; Moody, B.; Szolovits, P.; Celi, L.~A.; and Mark, R.~G.
\newblock 2016.
\newblock Mimic-iii, a freely accessible critical care database.
\newblock {\em Scientific data}.

\bibitem[\protect\citeauthoryear{Kolen and Pollack}{1991}]{kolen1991back}
Kolen, J.~F., and Pollack, J.~B.
\newblock 1991.
\newblock Back propagation is sensitive to initial conditions.
\newblock In {\em Advances in neural information processing systems}.

\bibitem[\protect\citeauthoryear{Krawczyk \bgroup et al\mbox.\egroup
  }{2017}]{krawczyk2017ensemble}
Krawczyk, B.; Minku, L.~L.; Gama, J.; Stefanowski, J.; and Wo{\'z}niak, M.
\newblock 2017.
\newblock Ensemble learning for data stream analysis: A survey.
\newblock {\em Information Fusion}.

\bibitem[\protect\citeauthoryear{Krogh and Vedelsby}{1995}]{krogh1995neural}
Krogh, A., and Vedelsby, J.
\newblock 1995.
\newblock Neural network ensembles, cross validation, and active learning.
\newblock In {\em Advances in neural information processing systems}.

\bibitem[\protect\citeauthoryear{Kulesza, Taskar, and
  others}{2012}]{kulesza2012determinantal}
Kulesza, A.; Taskar, B.; et~al.
\newblock 2012.
\newblock Determinantal point processes for machine learning.
\newblock {\em Foundations and Trends{\textregistered} in Machine Learning}.

\bibitem[\protect\citeauthoryear{Kuncheva and
  Whitaker}{2003}]{kuncheva2003measures}
Kuncheva, L.~I., and Whitaker, C.~J.
\newblock 2003.
\newblock Measures of diversity in classifier ensembles and their relationship
  with the ensemble accuracy.
\newblock {\em Machine learning} 51(2).

\bibitem[\protect\citeauthoryear{Liang}{2018}]{percytalk}
Liang, P.
\newblock 2018.
\newblock How should we evaluate machine learning for ai?
\newblock Thirty-Second AAAI Conference on Artificial Intelligence.

\bibitem[\protect\citeauthoryear{Lichman}{2013}]{uci}
Lichman, M.
\newblock 2013.
\newblock {UCI} ml repository.

\bibitem[\protect\citeauthoryear{Liu and Yao}{1999}]{liu1999simultaneous}
Liu, Y., and Yao, X.
\newblock 1999.
\newblock Simultaneous training of negatively correlated neural networks in an
  ensemble.
\newblock {\em IEEE Transactions on Systems, Man, and Cybernetics, Part B
  (Cybernetics)} 29(6).

\bibitem[\protect\citeauthoryear{Lorena \bgroup et al\mbox.\egroup
  }{2018}]{lorena2018complex}
Lorena, A.~C.; Garcia, L.~P.; Lehmann, J.; Souto, M.~C.; and Ho, T.~K.
\newblock 2018.
\newblock How complex is your classification problem? a survey on measuring
  classification complexity.
\newblock {\em arXiv preprint arXiv:1808.03591}.

\bibitem[\protect\citeauthoryear{Madry \bgroup et al\mbox.\egroup
  }{2017}]{madry2017towards}
Madry, A.; Makelov, A.; Schmidt, L.; Tsipras, D.; and Vladu, A.
\newblock 2017.
\newblock Towards deep learning models resistant to adversarial attacks.
\newblock {\em arXiv preprint arXiv:1706.06083}.

\bibitem[\protect\citeauthoryear{Pang \bgroup et al\mbox.\egroup
  }{2019}]{pang2019improving}
Pang, T.; Xu, K.; Du, C.; Chen, N.; and Zhu, J.
\newblock 2019.
\newblock Improving adversarial robustness via promoting ensemble diversity.
\newblock {\em arXiv preprint arXiv:1901.08846}.

\bibitem[\protect\citeauthoryear{Parascandolo \bgroup et al\mbox.\egroup
  }{2017}]{parascandolo2017learning}
Parascandolo, G.; Kilbertus, N.; Rojas-Carulla, M.; and Sch{\"o}lkopf, B.
\newblock 2017.
\newblock Learning independent causal mechanisms.
\newblock {\em arXiv preprint arXiv:1712.00961}.

\bibitem[\protect\citeauthoryear{Quionero-Candela \bgroup et al\mbox.\egroup
  }{2009}]{sugiyama2017dataset}
Quionero-Candela, J.; Sugiyama, M.; Schwaighofer, A.; and Lawrence, N.~D.
\newblock 2009.
\newblock {\em Dataset shift in machine learning}.

\bibitem[\protect\citeauthoryear{Ribeiro, Singh, and
  Guestrin}{2016}]{ribeiro2016should}
Ribeiro, M.~T.; Singh, S.; and Guestrin, C.
\newblock 2016.
\newblock Why should i trust you?: Explaining the predictions of any
  classifier.
\newblock In {\em Proceedings of the 22nd ACM SIGKDD International Conference
  on knowledge discovery and data mining}.
\newblock ACM.

\bibitem[\protect\citeauthoryear{Ross, Hughes, and
  Doshi-Velez}{2017}]{ross2017right}
Ross, A.~S.; Hughes, M.~C.; and Doshi-Velez, F.
\newblock 2017.
\newblock Right for the right reasons: Training differentiable models by
  constraining their explanations.
\newblock {\em arXiv preprint arXiv:1703.03717}.

\bibitem[\protect\citeauthoryear{Ross, Pan, and
  Doshi-Velez}{2018}]{ross2018learning}
Ross, A.; Pan, W.; and Doshi-Velez, F.
\newblock 2018.
\newblock Learning qualitatively diverse and interpretable rules for
  classification.
\newblock In {\em 2018 ICML Workshop on Human Interpretability in Machine
  Learning}.

\bibitem[\protect\citeauthoryear{Schapire}{1990}]{schapire1990strength}
Schapire, R.~E.
\newblock 1990.
\newblock The strength of weak learnability.
\newblock {\em Machine learning} 5(2).

\bibitem[\protect\citeauthoryear{Semenova and Rudin}{2019}]{semenova2019study}
Semenova, L., and Rudin, C.
\newblock 2019.
\newblock A study in rashomon curves and volumes: A new perspective on
  generalization and model simplicity in machine learning.
\newblock {\em arXiv preprint arXiv:1908.01755}.

\bibitem[\protect\citeauthoryear{Shoham and Permuter}{2019}]{shoham2019amended}
Shoham, R., and Permuter, H.
\newblock 2019.
\newblock Amended cross-entropy cost: An approach for encouraging diversity in
  classification ensemble (brief announcement).
\newblock In {\em International Symposium on Cyber Security Cryptography and
  Machine Learning}.

\bibitem[\protect\citeauthoryear{Szegedy \bgroup et al\mbox.\egroup
  }{2013}]{szegedy2013intriguing}
Szegedy, C.; Zaremba, W.; Sutskever, I.; Bruna, J.; Erhan, D.; Goodfellow, I.;
  and Fergus, R.
\newblock 2013.
\newblock Intriguing properties of neural networks.
\newblock {\em arXiv preprint arXiv:1312.6199}.

\bibitem[\protect\citeauthoryear{Tramèr \bgroup et al\mbox.\egroup
  }{2018}]{tramer2017ensemble}
Tramèr, F.; Kurakin, A.; Papernot, N.; Goodfellow, I.; Boneh, D.; and
  McDaniel, P.
\newblock 2018.
\newblock Ensemble adversarial training: Attacks and defenses.
\newblock In {\em International Conference on Learning Representations}.

\bibitem[\protect\citeauthoryear{Zadrozny}{2004}]{zadrozny2004learning}
Zadrozny, B.
\newblock 2004.
\newblock Learning and evaluating classifiers under sample selection bias.
\newblock In {\em Twenty-first International Conference on Machine learning}.

\bibitem[\protect\citeauthoryear{Zhou, Wang, and Bilmes}{2018}]{NIPS2018_7831}
Zhou, T.; Wang, S.; and Bilmes, J.~A.
\newblock 2018.
\newblock Diverse ensemble evolution: Curriculum data-model marriage.
\newblock In {\em Advances in Neural Information Processing Systems 31}.

\end{thebibliography}
\clearpage
\appendix

\section{Appendix}
\subsection{Imposing Penalties over Manifolds}\label{SUPP-sec:manifolds}

\begin{figure}[htb!]
  \centering
  \includegraphics[width=\columnwidth]{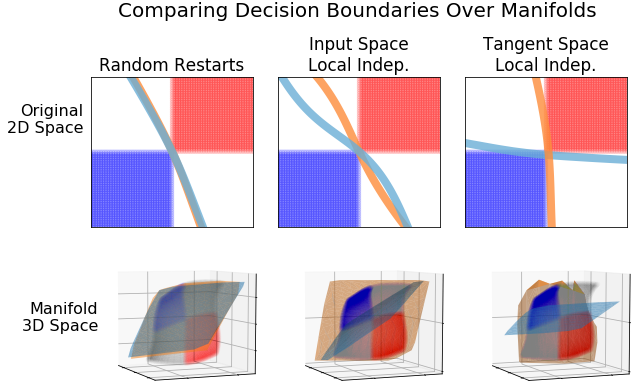}
  \caption{Synthetic 2D manifold dataset (randomly sampled from a neural network) embedded in $\mathbb{R}^3$, with decision boundaries shown in 2D chart
  space (top) and the 3D embedded manifold space (bottom). Naively imposing
  LIT penalties in $\mathbb{R}^3$ (middle) leads
  to only slight differences in the chart space decision boundary, but given
  knowledge of the manifold's tangent vectors (right), we can
  recover maximally different chart space boundaries.}
  \label{SUPP-fig:basic-manifold}
\end{figure}

In the beginning of our derivation of $\mathtt{CosIndepErr}$ (Equation~\ref{eq:gradpush}), we assumed that locally, $N_{\epsilon}(x) \approx \mathcal{B}_{\epsilon}(x)$. However, in many cases, our data manifold $\Omega_{x}$ is much lower dimensional than $\mathbb{R}^D$. In these cases, we have additional degrees of freedom to learn decision boundaries that, while locally orthogonal, are functionally equivalent over the dimensions which matter. To restrict spurious similarity, we can project our
gradients down to the data manifold. Given a local basis for its tangent space, we
can accomplish this by taking dot products between $\nabla f$ and $\nabla g$ and
each tangent vector, and then use these two vectors of dot products to compute
the cosine similarity in Equation~\ref{eq:cosgrad}. More formally, if $J(x)$ is the Jacobian
matrix of manifold tangents at $x$, we can replace our regular cosine penalty with
\begin{equation}
\begin{split}
  \mathtt{ManifIndepErr}(f,g) &\equiv \mathbb{E}\left[\cos^2(\nabla f\|_{J}(x), \nabla g\|_{J}(x)), \right]\\
   \text{where} \quad&\nabla f\|_{J}(x) = \nabla f(x)^\intercal J(x), \\
   &\nabla g\|_{J}(x) = \nabla g(x)^\intercal J(x)
   \end{split}
\end{equation}
An example of this method applied to a toy example is given in Figure \ref{SUPP-fig:basic-manifold}.
Alternatively, if we are using projected gradient descent adversarial training to
minimize the original formulation in Equation~\ref{eq:adverror}, we can modify
its inner optimization procedure to project input gradient updates back to the manifold.

For many problems of interest, we do not have a closed form expression for the
data manifold or its tangent vectors. In this case, however, we can approximate
one, e.g. by performing PCA or training an autoencoder. Local independence training can also simply be used
on top of this learned representation directly.

\subsection{Additional Figures}\label{SUPP-sec:more-mort}

\begin{figure}[htb!]
  \centering
  \includegraphics[width=\columnwidth]{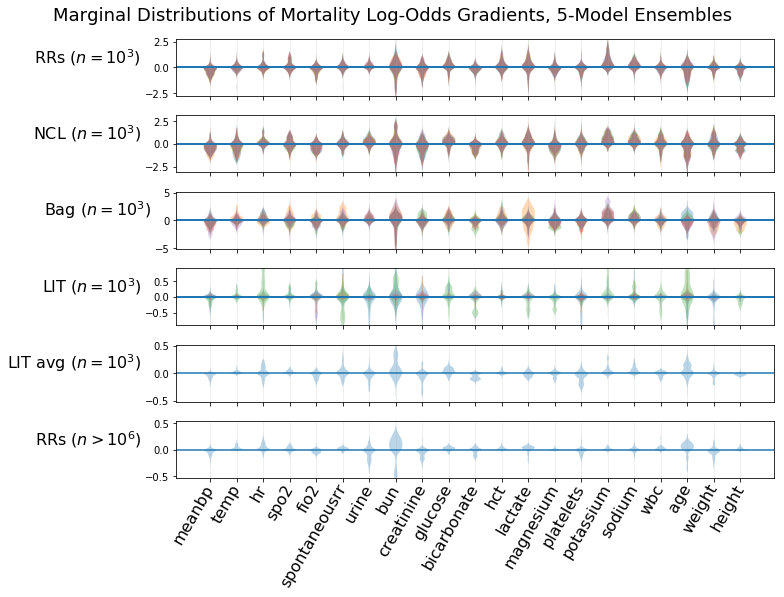}
  \caption{Violin plots showing marginal distributions of ICU mortality input gradients across heldout data for 5-model ensembles trained on the $n=1000$ slice (top 5 plots) and restarts on the full dataset (bottom). Distributions for each model in each ensemble are overlaid with transparency in the top 4 plots.
  From the top, we see that restarts and NCL learn models with similar gradient distributions. Bagging is slightly more varied, but only LIT (which performs significantly better on the prediction task) exhibits significant differences between models. When LIT gradients on this limited data task are averaged (second from bottom), their distribution comes to resemble (in both shape and scale) that of a model trained on the full dataset (bottom), which may explain LIT's stronger performance.}
  \label{SUPP-fig:extrap-grads}
\end{figure}

\begin{figure}[htb!]
  \centering
  \includegraphics[width=\columnwidth]{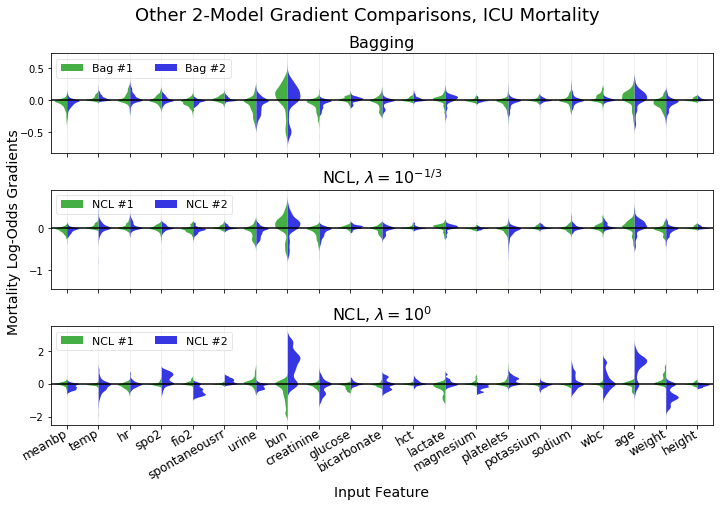}
  \caption{Companion to Figure~\ref{fig:sepsis-gradient-spectra} showing differences in the distributions of input gradients for other 2-model ensemble methods. Bagging is largely identical to random restarts, while NCL exhibits a sharp transition with $\lambda$.}
  \label{SUPP-fig:twomod-grads}
\end{figure}

\begin{figure*}[htb!]
  \includegraphics[width=2\columnwidth]{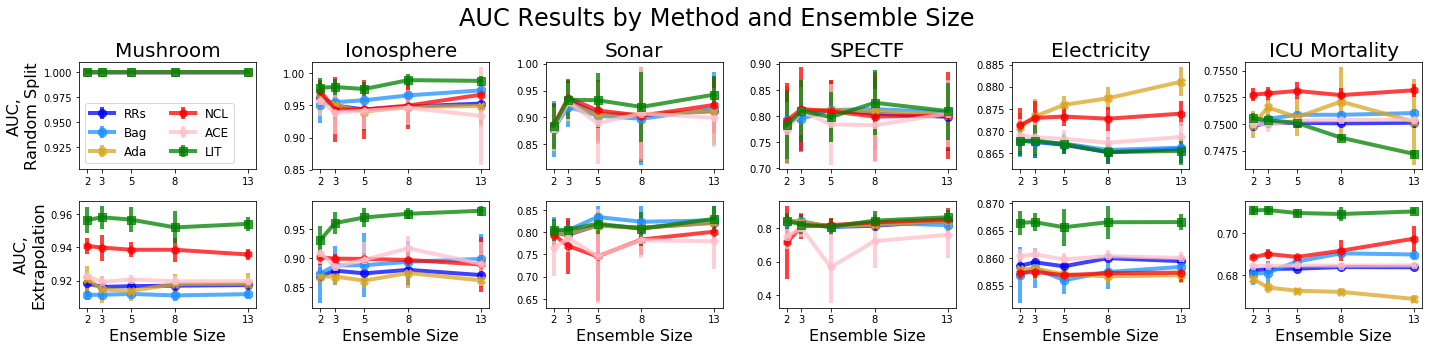}

  \vspace{0.5cm}

  \includegraphics[width=2\columnwidth]{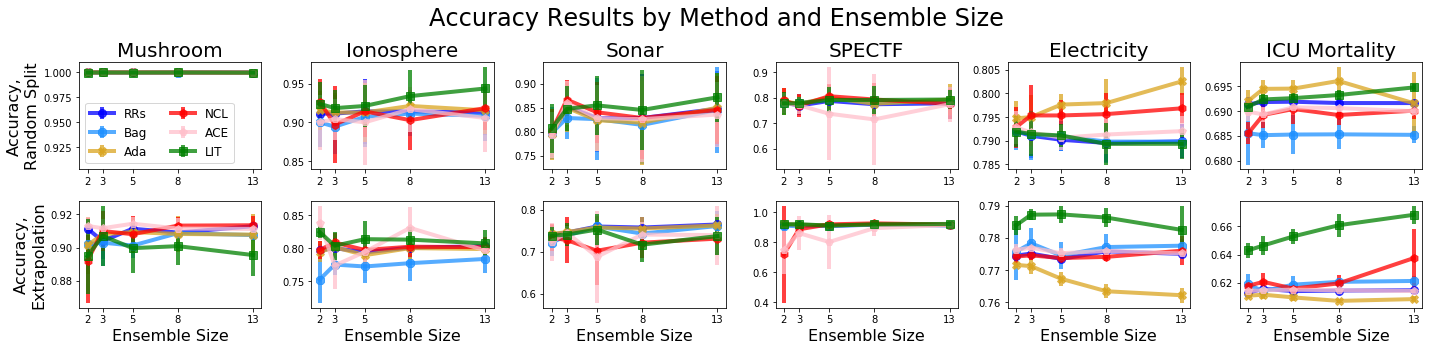}
  \caption{Full ensemble AUC and accuracy results by method and ensemble size. LIT usually beats baselines when train $\neq$ test, but the optimal ensemble size (cross-validated in the result tables in the main paper, but expanded here) can vary.}
  \label{SUPP-fig:all-aucs}
\end{figure*}

\begin{figure*}[htb!]
  \includegraphics[width=2\columnwidth]{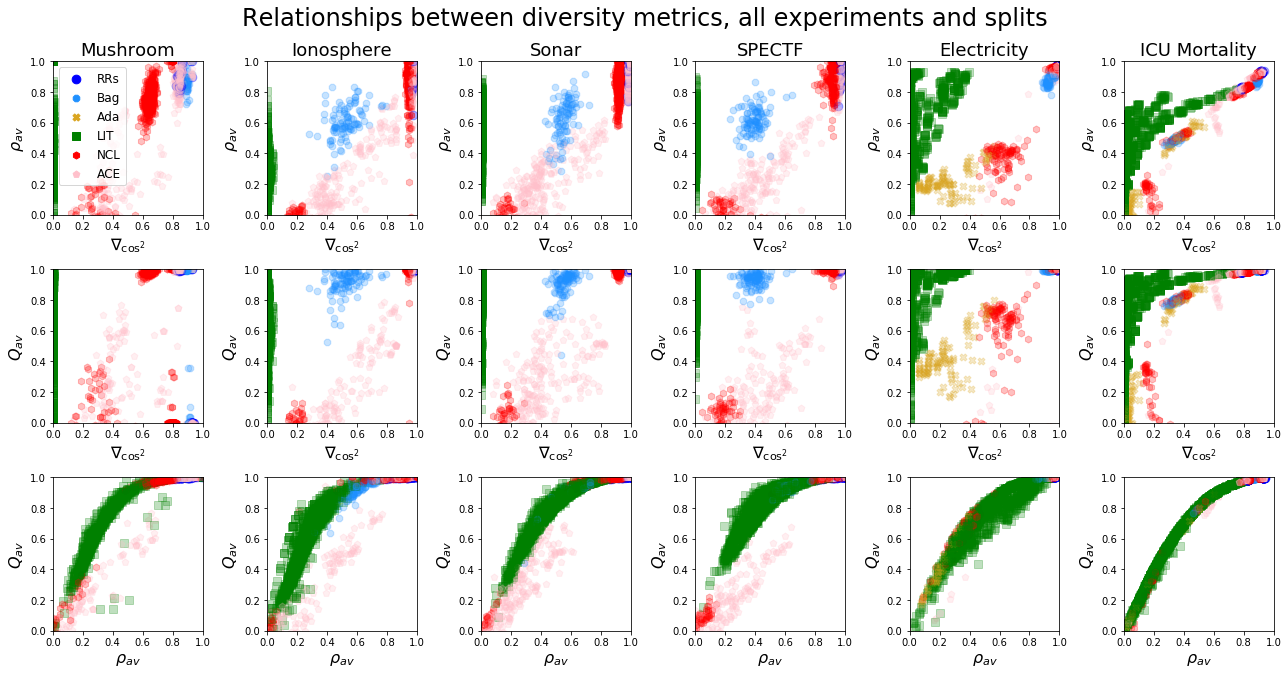}
  \caption{Empirical relationship between our similarity metric (or penalty) $\nabla_{\cos^2}$ and more classic measures of prediction similarity such as error correlation ($\rho_{av}$) and the Q-statistic ($Q_{av}$), with one marker for every method, $\lambda$, dataset, split, ensemble size, and restart. In general, we find meaningful relationships between $\nabla_{\cos^2}$ and classic diversity metrics, despite the fact that $\nabla_{\cos^2}$ does not require ground-truth labels. The bottom row of this figure also shows that LIT models (green) tend to have lower and more widely varying $Q_{av}$ and $\rho_{av}$, indicating greater ability to control heldout prediction diversity through training $\lambda$. We also measured the interrater agreement $\kappa$ but we found the results almost identical to $\rho_{av}$ and omit them to save space.}
  \label{SUPP-fig:rho-vs-cos}
\end{figure*}

\subsection{Orthogonality and Independence}\label{SUPP-sec:loc-indep}

Here we briefly explain the alternate statistical formulation of local independence given in Section~\ref{sec:diverse}.
Let $x \in \mathbb{R}^D$ be a data point, let $f_1$ and $f_2$ be functions from $\mathbb{R}^D \to \mathbb{R}$, and let $\randeps \sim \mathcal{N}(0, \sigma^2\mathbbm{1})$ be a small perturbation. Taylor expanding $f_1$ and $f_2$ around $x$ and assuming all gradients are nonzero at $x$, we have that \[
\begin{split}
  f_i(x + \randeps) &\approx f_i(x) + \nabla f_i(x) \cdot \randeps + \mathcal{O}(\randeps\cdot\randeps) \\
\end{split}
\]
Defining the changes in each $f_i$ under the perturbation of $x$ to be $\Delta f_i(x)$, and assuming that all gradients are nonzero at $x$ and that we select $\sigma^2$ to be small enough that $||\randeps||^2 \ll ||\randeps||$ with high probability, we have \[
  \Delta f_i(x) \equiv f_i(x + \randeps) - f_i(x) \approx \nabla f_i(x)\cdot\randeps
\]
If we now consider $\mathrm{Cov}(\Delta f_1(x), \Delta f_2(x))$, the covariance of the changes in each function (which we term the ``local covariance'' between $f_1$ and $f_2$ at $x$), we see that \[
\begin{split}
  \mathrm{Cov}&(\Delta f_1(x), \Delta f_2(x)) \\
  &\approx \mathbb{E}[(\nabla f_1(x)\cdot  \randeps - \mathbb{E}[\nabla f_1(x)\cdot  \randeps]) \\ 
  &\hspace{0.8cm}(\nabla f_2(x)\cdot  \randeps -\mathbb{E}[\nabla f_2(x)\cdot  \randeps])] \\
  &= \mathbb{E}[(\nabla f_1(x)\cdot  \randeps)(\nabla f_2(x)\cdot  \randeps)] \\
  &= \mathbb{E}[ \randeps \cdot  \randeps ] \nabla f_1(x)\cdot \nabla f_2(x) \\
  &= \sigma^2 \nabla f_1(x)\cdot  \nabla f_2(x) \\
\end{split}
\]
Normalizing this quantity, we see the local correlation is \[
  \begin{split}
    \mathrm{Corr}(\Delta f_1(x), \Delta f_2(x)) &= \frac{\nabla f_1(x)\cdot  \nabla f_2(x)}{||\nabla f_1(x)||_2 ||\nabla f_2(x)||_2} \\
    &= \cos(\nabla f_1(x), \nabla f_2(x)).
  \end{split}
\]
Finally, let's consider the distribution of $\Delta f_i(x)$. Since $\Delta f_i(x)$ is approximately equal to $\nabla f_i(x)\cdot \randeps$, which is a dot product between a deterministic vector ($\nabla f_i(x)$) and a Gaussian sample ($\randeps$), then $\Delta f_1(x)$ and $\Delta f_2(x)$ are approximately equal to sums of Gaussian random variables and are therefore themselves Gaussian. Noting that the entropy of Gaussians with covariance matrices $\Sigma$ is $H(X)=\frac{1}{2}\ln(2\pi e |\Sigma|)$ and for a bivariate Gaussian random variable $(X,Y)$, its covariance determinant $|\Sigma| = \mathrm{Var}(X)\mathrm{Var}(Y)-\mathrm{Cov}(X,Y)^2$,
\[
\begin{split}
  I(X,Y)
    &= H(X) + H(Y) - H(X,Y) \\
    &= \frac{1}{2}\ln \left( \frac{\mathrm{Var}(X) \mathrm{Var}(Y)}{\mathrm{Var}(X)\mathrm{Var}(Y) - \mathrm{Cov}(X,Y)^2} \right) \\
    &= -\frac{1}{2}\ln \left( 1 -  \frac{\mathrm{Cov}(X,Y)^2}{\mathrm{Var}(X) \mathrm{Var}(Y)} \right) \\
    &= -\frac{1}{2}\ln \left( 1 - \mathrm{Corr}(X,Y)^2 \right).
\end{split}
\]
So, between two 1D Gaussians, zero correlation implies statistical independence. Therefore, if $f_1$ and $f_2$ have orthogonal and nonzero input gradients at $x$, their changes under small $\mathcal{N}(0, \sigma^2\mathbbm{1})$ perturbations are independent.

\end{document}